\documentclass[twoside,11pt]{article}
\usepackage{proceed2e}
\usepackage{ifthen}
\usepackage{comment}
\usepackage{pgf,tikz}
\usepackage{amsmath,amssymb}
\usepackage{stmaryrd}
\usepackage{draftcopy}
\usepackage{multirow}
\usepackage{subfig}

\newcounter{isamac} 
\ifthenelse{\value{isamac}=1}{
\input BoxedEPS          %
\SetTexturesEPSFSpecial  
\HideDisplacementBoxes   %
}{}

\newtheorem{theorem}{Theorem}

\newtheorem{lemma}[theorem]{Lemma}
\newtheorem{definition}[theorem]{Definition}

\newcommand{\indm}[2]{\ensuremath{{\mathfrak I}_{\kern-1pt\scriptstyle#1}({\mathcal
#2})}} 

\newcommand{\ind}{\mbox{$\perp \kern-5.5pt \perp$}}

\newcommand{\uned}{\hbox{\kern3pt\raise2.5pt\vbox{\hrule
width9pt height 0.3pt}\kern3pt}}

\newcommand{\dashed}{\hbox{\kern3.05pt\raise2.5pt\vbox{\hrule
width1.7pt height 0.3pt}\kern1.8pt\raise2.5pt\vbox{\hrule
width1.7pt height 0.3pt}\kern1.8pt\raise2.5pt\vbox{\hrule
width1.7pt height 0.3pt}\kern1.8pt\raise2.5pt\vbox{\hrule
width1.7pt height 0.3pt}\kern3.05pt}}

\newcommand{\lhead}{\ensuremath{\prec}}

\newcommand{\head}{\ensuremath{\succ}}

\newcommand{\pedg}[2]{\ensuremath{{\kern0.5pt
\scriptstyle{\ifthenelse{\equal{\head}{#1}}{\lhead\kern0.5pt}{#1\kern0.5pt}}\joinrel\relbar
\negthinspace\relbar\joinrel{\kern0.5pt #2}\kern0.5pt}}}

\newcommand{\pdots}{\hbox{\kern2.5pt\raise1.5pt\hbox{\ensuremath{\ldots}}\ke
rn2.5pt}}  

\def\ci{\perp\!\!\!\perp}

\DeclareMathOperator{\dis}{dis}

\DeclareMathOperator{\pa}{pa}
\DeclareMathOperator{\de}{de}
\DeclareMathOperator{\spo}{sp}

\DeclareMathOperator{\an}{an}
\DeclareMathOperator{\mb}{mb}

\DeclareMathOperator{\tailo}{tail}

\begin{document}


\title{Parameter and Structure Learning in Nested Markov Models}

\author{
Ilya Shpitser \\
\And
Thomas S. Richardson \\
\And
James M. Robins \\
\And
Robin Evans\\
}


\maketitle

\begin{abstract}%
The constraints arising from DAG models with latent variables can be
naturally represented by means of acyclic directed mixed graphs (ADMGs).
Such graphs contain directed ($\to$) and bidirected $(\leftrightarrow)$
arrows, and contain no directed cycles.  DAGs with latent variables imply
independence constraints in the distribution resulting from a `fixing'
operation, in which a joint distribution is divided by a conditional. This
operation generalizes marginalizing and conditioning. Some of these
constraints correspond to identifiable `dormant' independence
constraints, with the well known `Verma constraint' as one example.
Recently, models defined by a set of the constraints arising after fixing
from a DAG with latents, were characterized via a recursive factorization
and a nested Markov property. In addition, a parameterization was given in
the discrete case.  In this paper we use this parameterization to describe
a parameter fitting algorithm, and a search and score structure learning
algorithm for these nested Markov models. We apply our algorithms to a
variety of datasets.
\end{abstract}


\section{Introduction}


Many data-generating process correspond to distributions that
factorize according to a directed acyclic graph (DAG).
Such models also have an intuitive
causal interpretation:  an arrow from a variable $X$ to a variable $Y$ in a
DAG model can be interpreted, in a way that can be made precise, to mean
that $X$ is a ``direct cause'' of $Y$.

In many contexts we do not observe all of the variables in the
data-generating process.
This creates major challenges for structure learning and for identifying
causal intervention distributions.
   While existing machinery based on DAGs with latent variables can
be applied to such settings, this creates a number of problems.  First,
there will in general be an infinite number of DAG models such that a
particular
margin of these models may represent the observed distribution; this is
still true if we
require the graph to be faithful.  Second,
prior knowledge about latent variables is often scarce, which implies any
modeling assumptions made by explicitly representing latents leaves one
open
to model misspecification bias.  An alternative approach, is to consider
  graphical models represented by  graphs
containing directed and bidirected edges, called Acyclic Directed Mixed
Graphs (ADMGs). In a companion paper we define
a `nested' Markov property for ADMGs that encodes independence constraints under a
`fixing' operation that divides the joint distribution by a conditional
density.
Given a DAG $\cal G$ with latent variables there is an ADMG ${\cal G}^*$
naturally associated with $\cal G$ via the operation of `latent
projection' \cite{verma90equiv}; the vertices of ${\cal G}^*$ are solely
the subset of vertices in $\cal G$ that are observed. 
We show that the observed distribution resulting from the DAG with latent
variables $\cal G$ obeys the nested Markov property associated with the
corresponding latent projection ${\cal G}^*$.

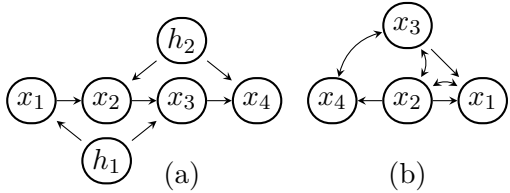
\begin{figure}
\begin{center}
  \begin{tikzpicture}[ >=stealth, node distance=1.0cm]
  \begin{scope}
    \tikzstyle{format} = [draw, thick, rectangle, minimum size=6mm, rounded
        corners=3mm]
    \path[->]	node[format] (x1) {$x_1$}
    		node[format, right of=x1] (x2) {$x_2$}
                  (x1) edge (x2)
		node[format, right of=x2] (x3) {$x_3$}
                  (x2) edge (x3)
		node[format, right of=x3] (x4) {$x_4$}
                  (x3) edge (x4)
		node[format, above of=x3, yshift=-0.2cm] (h2) {$h_2$}
		node[format, below of=x2, yshift=+0.2cm] (h1) {$h_1$}
		  (h1) edge (x1)
		  (h1) edge (x3)
		  (h2) edge (x2)
		  (h2) edge (x4)

	node[below of=x3] {(a)}
	;
	
  \end{scope}
  \begin{scope}[xshift=4.0cm]
    \tikzstyle{format} = [draw, thick, rectangle, minimum size=6mm, rounded
        corners=3mm,color=black]
    \path[->]	node[format] (x4) {$x_4$}
    		node[format, right of=x4] (x2) {$x_2$}
                  (x2) edge (x4)
		node[format, right of=x2] (x1) {$x_1$}
		  (x2) edge (x1)
		  (x2) edge[<->, bend left] (x1)
		node[format, above of=x2] (x3) {$x_3$}
		  (x3) edge (x1)
		  (x3) edge[<->, bend left] (x2)
		  (x3) edge[<->, bend right] (x4)
	node[below of=x2] {(b)}
	;
  \end{scope}

  \end{tikzpicture}
\end{center}
\caption{(a) A latent variable DAG not entailing any d-separation statements on
$x_1,x_2,x_3,x_4$.
(b) An ADMG $\mathcal{G}$ with missing edges representing a saturated
nested Markov model.}
\label{verma1}
\end{figure}

Previous work \cite{evans10maximum} has given a discrete parameterization and
ML fitting algorithms, as well as a characterization of model equivalence for
mixed graph models representing conditional independences \cite{ali09equiv}.
It is well-known, however, that
models representing DAG marginals can contain non-parametric constraints
which cannot be represented as conditional independence constraints.
For instance, in any density $P$ Markov relative to a DAG with latents represented by
the graph shown in Fig. \ref{verma1} (a),
it is known (see \cite{verma90equiv,robins86new}) that
\begin{equation}\label{eq:verma}
\frac{\partial}{\partial x_1} \sum_{x_2} P(x_4 \mid x_1,x_2,x_3) P(x_2 \mid x_1) = 0 
\end{equation}
This constraint can be viewed as stating that $X_4$ is independent of
$X_1$ in the distribution obtained from $P(x_1,x_2,x_3,x_4)$ after dividing
by a conditional $P(x_3|x_2,x_1)$ \cite{robins:1999}.
Also note that the expression
(\ref{eq:verma}) is an instance of the g-formula of \cite{robins86new}.
If the graph shown in Fig. \ref{verma1} (a)
is causal, then this constraint can be interpreted as an (identifiable)
dormant independence constraint \cite{shpitser08dormant}, which states that
$X_4$ is independent of $X_1$ given $do(x_3)$, where $do(.)$ denotes
an intervention \cite{pearl00causality}.

Since the DAG in Fig. \ref{verma1} (a) implies no conditional independences
on the 4 observable variables, the set of marginal distributions on $x_1,x_2,x_3,x_4$
obtained from densities over $x_1,x_2,x_3,x_4,h_1,h_2$ Markov relative to this DAG
is a saturated model when viewed as a 
model of conditional independence.  Thus, any structure learning
algorithm which only relies on conditional independence constraints will
return a (maximally uninformative) unoriented complete graph when given data sampled from
one of such marginal distributions.

Nevertheless, it is possible to use constraints such as (1), which we call
post-truncation independences or reweighted independences to distinguish
between models, with appropriate assumptions.  In \cite{shpitser09edge},
such constraints
were used to test for the presence of certain direct causal effects (represented
by directed arrows in graphical causal models).  A recent paper
\cite{richardson12nested}
has given a nested factorization for mixed graph models which implies,
along with
the standard conditional independences in mixed graphs, post-truncation
independences of the type shown in (1).  Furthermore, a parameterization for
discrete models based on this factorization was given.  Another recent paper
\cite{shpitser11eid} has taken advantage of this parameterization to give
a general algorithm for efficiently computing causal effects.
In this paper, we take advantage of this parameterization to give a maximum
likelihood parameter fitting algorithm for mixed graph models of
post-truncation independence.  Furthermore, we use this algorithm to
construct a search and score algorithm based on BIC \cite{schwarz78bic} for
learning
mixed graph structure while taking advantage of post-truncation independences.

The paper is organized as follows.  In section 2, we introduce the graphical,
and probabilistic preliminaries necessary for the remainder of the
paper.  In section 3, we introduce nested Markov models.
 In section 4, we give a parameterization of discrete nested Markov models.
In section 5, we describe the parameter fitting algorithm.  In section 6,
we describe the search and score algorithm.
Section 7 contains our experiments.  Section 8 gives our conjecture for the
characterization of equivalence classes of mixed graphs over four nodes, and
gives experimental evidence in favor of our conjecture.  Section 9 contains
the discussion and concluding remarks.

\section{Preliminaries}

A \emph{directed mixed graph} $\mathcal{G}(V,E)$ is a graph with a set of
vertices $V$ and a set of edges $E$ which may contain directed
($\to$) and bidirected $(\leftrightarrow)$ edges.  A directed cycle is a
path of the form $x \to \cdots \to y$ along with an edge $y \to x$.  An
\emph{acyclic} directed mixed graph (ADMG) is a mixed graph containing no
directed cycles.

\subsection{Conditional ADMGs}

A {\it conditional} acyclic directed mixed graph (CADMG) ${\cal G}(V,W,E)$ is
an ADMG with a vertex set
$V\cup W$, where $V\cap W = \emptyset$, subject to the restriction that for all $w \in W$, $\pa_{\cal G}(w) = \emptyset = \spo_{\cal G}(w)$.

Whereas an ADMG with vertex set $V$ represents a joint density $p(x_V)$, a conditional ADMG
is a graph with two disjoint sets of vertices, $V$ and $W$ that is used to represent the Markov structure of a {`kernel'} $q_V(x_V| x_W)$.
Following \cite[p.46]{lauritzen96graphical}, we define a {\em kernel} to be a non-negative function $q_V(x_V| x_W)$ satisfying:
\begin{equation}\label{eq:cond}
\sum_{x_V \in {\mathfrak{X}_V}} q_V(x_V \mid x_W) = 1\quad\quad \hbox{for all }x_W \in \mathfrak{X}_W.
\end{equation}
We use the term `kernel' and write $q_V(\cdot | \cdot)$ (rather than $p(\cdot | \cdot)$)  to emphasize that these functions,
though they satisfy (\ref{eq:cond}) and thus most properties of conditional densities, will not, in general, be formed via
the usual operation of conditioning on the event $X_W=x_W$.
To conform with standard notation for densities, we define for every $A \subseteq V$,
$q_V(x_A | x_W) \equiv \sum_{V \setminus A} q_V(x_V | x_W)$, and
$q_V(x_{V \setminus A} | x_{W \cup A}) \equiv \frac{q_V(x_V | x_W)}{q_V(x_A | x_W)}$.

For a CADMG ${\cal G}(V,W,E)$ we consider collections 
of random variables $(X_{v})_{ v\in V}$ taking values in probability 
spaces $({\mathfrak X}_{v})_{v\in V}$ conditional on variables
 $(X_{w})_{ w\in W}$ with state spaces $({\mathfrak X}_{w})_{w\in W}$.
In all the cases we consider the probability spaces 
are either real finite-dimensional vector spaces or finite discrete 
sets. For $A\subseteq V\cup W$ we let ${\mathfrak X}_{A}\equiv \times_{u\in A}
({\mathfrak X}_{u})$, and $X_{A}\equiv(X_{v})_{v\in 
A}$. We use the usual shorthand notation: $v$ denotes a 
vertex and a random variable $X_{v}$, likewise $A$ 
denotes a vertex set and $X_{A}$.
It is because we will always condition on the variables in $W$ that
we do not permit edges between vertices in $W$.

An ADMG ${\cal G}(V,E)$ may be seen as a CADMG in which $W=\emptyset$.
In this manner, though we will state subsequent definitions for CADMGs, they will also apply  to ADMGs.


The induced subgraph of  a CADMG ${\cal G}(V,W,E)$ given by set $A$, denoted ${\cal G}_A$ consists of 
${\cal G}(V\cap A,W\cap A,E_{A})$, where $E_{A}$ is the set of edges in $\cal G$ with both endpoints in $A$.
Note that in forming ${\cal G}_A$ , the status of the vertices in $A$ with regard to whether they are in $V$ or $W$ is preserved.

\begin{comment}
\begin{figure}
\begin{center}
  \begin{tikzpicture}[>=stealth, node distance=1.2cm]

  \begin{scope}
    \tikzstyle{format} = [draw, thick, rectangle, minimum size=6mm, rounded
        corners=3mm]
    \tikzstyle{square} = [draw, thick, rectangle, minimum size=5mm]
    \path[->]	node[square] (x1) {$x_1$}
    		node[format, right of=x1] (x2) {$x_2$}
                  (x1) edge (x2)
		node[square, right of=x2] (x3) {$x_3$}
		node[format, right of=x3] (x4) {$x_4$}
                  (x3) edge (x4)
                  (x2) edge[<->, bend left] (x4);

  \end{scope}

  \end{tikzpicture}
\end{center}
\caption{An example CADMG.}
\label{verma2}
\end{figure}
\end{comment}

\subsection{Districts}

A set $C$ is \emph{connected} in $\mathcal{G}$ if every pair of vertices in
$C$ are connected by a path with every vertex on the path contained
in $C$.  A connected set $C$ in an ADMG $\mathcal{G}$ is \emph{inclusion
maximal} if no superset of $C$ is connected.

For a given CADMG $\mathcal{G}(V,W,E)$, the \emph{induced bidirected graph}
$(\mathcal{G})_{\leftrightarrow}$ is the CADMG formed by removing all directed
edges from $\mathcal{G}$.  Similarly, $(\mathcal{G})_{\to}$ is formed by
removing all bidirected edges.  A set connected in
$(\mathcal{G})_{\leftrightarrow}$ is called bidirected connected.

For a given vertex $x \in V$ in $\mathcal{G}$, the district (c-component) of
$x$, denoted by $dis_{\mathcal{G}}(x)$ is the connected component of
$(\mathcal{G})_{\leftrightarrow}$.
Districts in an ADMG $\mathcal{G}(V,E)$ form a partition of $V$.
In a DAG $\mathcal{G}(V,E)$ the set of districts is the set of all single
element node sets of $V$.  In a CADMG, all districts are subsets of $V$, the
nodes of $W$ are not included by definition.
For an induced subgraph $\mathcal{G}_A$, we write
$dis_A(x)$ as a shorthand for $dis_{\mathcal{G}_A}(x)$.

\subsection{The fixing operation and fixable vertices}

We now introduce a `fixing' operation on an ADMG or CADMG that has the effect of transforming
a random vertex into a fixed vertex, thereby changing the graph. However, in general this operation may only be applied to a subset of the vertices in the graph, which we term the set of (potentially) fixable vertices.
\begin{definition}
Given a CADMG ${\mathcal{G}}(V,W)$ the set of {\em fixable vertices},
\[
{\mathbb F}({\mathcal{G}}) \equiv \left\{ v \mid v \in V, \dis_{\mathcal{G}}(v) \cap \de_{\mathcal{G}}(v) = \{v\}\right\}.
\]
\end{definition}
In words, a vertex $v$ is fixable in $\mathcal{G}$ if there is no vertex $v^*$ that is both a descendant of $v$ and
in the same district as $v$ in  $\mathcal{G}$.

\begin{definition}
Given a CADMG ${\mathcal{G}}(V,W,E)$, and a kernel $q_V(X_V \mid X_W)$,
for every $r \in {\mathbb F}({\mathcal{G}})$ we associate a {\em fixing transformation}
$\phi_r$ on the pair $({\mathcal{G}}, q_V(X_V \mid X_W))$ defined as follows:
\[
\phi_r ({\mathcal{G}}) \equiv
{\mathcal{G}}^*(V\setminus \{r\}, W \cup \{r\}, E_{r}),
\]
where
 $E_{r}$ is the subset of edges in $E$ that do not have arrowheads into $r$,
 and
\begin{equation}\label{eq:fix-kernel}
\phi_r (q_V(x_V \mid x_W); {\mathcal{G}})\equiv
\frac{q_V(x_V \mid x_W)}{q_V(x_r \mid x_{\mb_{\mathcal{G}}(r, \an_{\mathcal{G}}(\dis_{\mathcal{G}}(r)))})}.
\end{equation}
\end{definition}
We use $\circ$ to indicate composition of operations in the natural way, so that:
$\phi_{r}\circ \phi_{s} ({\mathcal{G}}) \equiv \phi_{r}( \phi_{s} ({\mathcal{G}}))$ and
$\phi_{r}\circ \phi_{s} (q_V(X_V | X_W); {\mathcal{G}}) \equiv
\phi_{r}\left(\phi_s\left(q_V(X_V | X_W); {\mathcal{G}}\right); \phi_s({\mathcal{G}})\right)$.
\subsection{Reachable and Intrinsic Sets}

In order to define our factorization, we will need to define special
classes of vertex sets in ADMGs.

\begin{definition}
A CADMG  ${\mathcal{G}}(V,W)$ is {\em reachable} from an ADMG ${\mathcal{G}}^*(V\cup W)$ if there is an ordering
of the vertices in $W=\langle w_1,\ldots , w_k\rangle$, such that
for $j=1,\ldots , k$,
\begin{align*}
w_1 \in {\mathbb F}({\mathcal{G}}^*) \hbox{ and  for }j= 2,\ldots, k,\\
w_j \in {\mathbb F}(\phi_{w_{j-1}}\circ \cdots \circ \phi_{w_{1}}({\mathcal{G}}^*)).
\end{align*}

\end{definition}
In words, a subgraph is reachable if, under some ordering, each of the vertices in $W$ may be fixed,
first in ${\mathcal{G}}^*$, and then in $\phi_{w_1}({\mathcal{G}}^*)$, then in  $\phi_{w_2}(\phi_{w_1}({\mathcal{G}}^*))$, and so on.
If a CADMG ${\mathcal{G}}(V,W)$ is reachable from ${\mathcal{G}}^*(V \cup W)$, we say that the set $V$ is reachable in ${\mathcal G}^*$.
Note that a reachable set $R$ in ${\mathcal G}$ may be obtained by fixing
vertices using more than one valid sequence.  We will denote
any valid composition of fixing operations that fixes a set $A$ by $\phi_A$ if applied to the graph, and by $\phi_{X_A}$ if applied to a kernel.
Note that with a slight abuse of notation (though justified as we will later see) we suppress the precise fixing sequence chosen.

\begin{definition} A set $C$ is {\em intrinsic} in $\mathcal{G}$ if it is a district in a reachable subgraph of $\mathcal{G}$. The set of intrinsic sets in an ADMG $\mathcal{G}$ is denoted by ${\cal I(\mathcal{G})}$.
\end{definition}

\begin{comment}
\begin{definition}
Let $C$ be a bidirected connected set in a CADMG $\mathcal{G}$.  Define
the following operations on subgraphs of $\mathcal{G}$:

\begin{eqnarray*}
\alpha_C:\quad  {\cal G} &\mapsto& {\cal G}_{an_{\cal G}(C)},\\
\delta_C:\quad {\cal G} &\mapsto& {\cal G}_{dis_{\cal G}(C)},\\
\gamma_C:\quad {\cal G} & \mapsto & \alpha_C(\delta_C({\cal G})),\\
\gamma_C^{(k)}:\quad {\cal G} & \mapsto &  \underbrace{\gamma_{C} (\cdots \gamma_{C} ({\cal G})\cdots )}_{k\hbox{\small -times}}.
\end{eqnarray*}

A bidirected connected set $C \subseteq V$ is intrinsic in $\mathcal{G}$
if for some
$k$, $\mathcal{G}_C = \gamma_C^{(k)}(\mathcal{G})$.  The set of all
intrinsic sets in a CADMG $\mathcal{G}$ will be denoted
$\mathcal{I}(\mathcal{G})$.
\end{definition}
\end{comment}

Note that in any DAG $\mathcal{G}(V,E)$,
$\mathcal{I}(\mathcal{G}) = \{ \{ x \} | x \in V \}$, while in any bidirected
graph $\mathcal{G}$, $\mathcal{I}(\mathcal{G})$ is equal to the set of all
connected sets in $\mathcal{G}$.

\begin{comment}
If a CAMDG represents a semi-Markovian model
(possibly after some intervention), then we can view the set
$\mathcal{I}(\mathcal{G})$ as the set of all connected sets $C$ such that
$P(c | do(pa(c)))$ is identifiable from $P(v)$ in $\mathcal{G}$.
The recursive definition of $\gamma_C$ mirrors the operations of the
identification algorithm used to identify causal effects in semi-Markovian
models \cite{tian02general},\cite{shpitser07hierarchy}.
\end{comment}

\section{Nested Markov Models}

We define a factorization on probability distributions represented by
ADMGs via intrinsic sets.
\begin{comment}
Before giving our factorization, we give the following definition.

\begin{definition}
For each district $D$ in a CAMDG $\mathcal{G}$ we define:

$$\mathcal{G}[D] \equiv \mathcal{G}^*(V^* = D, W^* = pa(D) \setminus D, E^*),$$

where edges between vertices in $V^*$ are edges in $\mathcal{G}_D$, no edge
between $W^*$ is an element of $E^*$, edges between vertices $w \in W^*$ and
$v \in V^*$ are those edges present in
$\mathcal{G}_{D \cup pa_{\mathcal{G}}(D)}$ which take the form $w \to v$.

\end{definition}

We now define our factorization.
\end{comment}

\begin{definition}[nested factorization] \label{def:factor-recurse}
Let ${\cal G}(V,E)$ be an ADMG.
A distribution $p(X_V)$ obeys the {\em nested factorization
according to  ${\cal G}(V,E)$}
if for every reachable subset $A \subseteq V$,
$\phi_{X_{V \setminus A}}(p(x_V); {\cal G}) = \prod_{D \in {\cal D}(\phi_{A}({\cal G}))} f_D(x_D | x_{\pa_{\cal G}(D) \setminus D})$.
\end{definition}

A distribution $p(x_V)$ that obeys the nested factorization with respect to ${\cal G}$ is said to be in the nested
Markov model of ${\cal G}$.

\begin{theorem}
If $p(x_V)$ is in the nested Markov model of ${\cal G}$, then for any reachable set $A$ in ${\cal G}$, any valid fixing
sequence on $V \setminus A$ gives the same CADMG over $A$, and the same kernel $q_A(x_A | x_{V \setminus A})$ obtained
from $p(x_V)$.
\end{theorem}

Due to this theorem, our decision to suppress the precise fixing sequence from the fixing operation is justified.

It is known that nested Markov factorization implies the global Markov property for ADMGs.

\begin{theorem}
If a distribution $p(X_V)$ is in the nested Markov model for ${\cal G}(V,E)$  then $p(X_V)$ obeys the global Markov
property for ${\cal G}(V,E)$.
\end{theorem}

The proof appears in \cite{richardson12nested}.
This result implies nested Markov models capture all conditional independence
statements normally associated with mixed graphs.  In addition, nested Markov
models capture additional independence constraints that manifest after
truncation operations.  For example, all distributions contained in the model
that factorizes according to the graph shown in Fig. \ref{verma1} (a), obey the
constraint that $X_1$ is independent of $X_4$ after ``truncating out''
(that is, dividing by) $P(x_3 \mid x_{2,1})$.

\section{Parameterization of Binary Nested Markov Models}

We now give a parameterization of nested Markov models.  The approach
generalizes in a straightforward way to finite discrete state spaces.

\subsection{Heads and Tails of Intrinsic Sets}


\begin{comment}
\begin{lemma}
Let $C$ be a bidirected connected set in a CADMG $\mathcal{G}$.  Then there
exists a unique minimal intrinsic set $int_{\mathcal{G}}(C)$ containing $C$.
\end{lemma}

This is shown in \cite{richardson11generalized}.

We refer to $int_{\mathcal{G}}(C)$ as the \emph{intrinsic closure} or
simply \emph{closure} of $C$.
\end{comment}

\begin{definition}
For an intrinsic set $C \in \mathcal{I}(\mathcal{G})$ of a CADMG $\mathcal{G}$,
define the recursive head (rh) as:
$rh(C) \equiv \{ x | x \in C; ch_{\mathcal{G}_C}(x) = \emptyset \}$.
\end{definition}

\begin{comment}
To invoke the causal interpretation of intrinsic sets, if a bidirected
connected set $C$
is not intrinsic in $\mathcal{G}$, then $P(c | do(pa(c)))$ is not identifiable
in a semi-Markovian model represented by $\mathcal{G}$, and moreover,
$C$ and $int_{\mathcal{G}}(C)$ form a hedge \cite{shpitser06idc} witnessing
this non-identifiability.
\end{comment}

\begin{definition}
The \emph{tail} associated with a recursive head $H$ of an intrinsic set $C$
in a CADMG $\mathcal{G}$ is given by:
$\tailo(H) \equiv (C \setminus H) \cup \pa_{\mathcal{G}}(C)$.
\end{definition}

\subsection{Binary Parameterization}

Multivariate binary distributions which obey the nested factorization with respect to an CADMG
$\mathcal{G}$ may be parameterized by the following parameters:

\begin{definition} The {\em binary parameters} associated with a CADMG $\cal G$
are a set of functions:
$\mathfrak{Q}_{\mathcal{G}} \equiv
\left\{ q_C(X_{H} = \textbf{0} | x_{\tailo(H)}) | H = rh(C), C \in {\cal I}({\cal G}) \right\}$.
\end{definition}
\noindent
Intuitively, a parameter $q_C(X_{H} = \textbf{0} | x_{\tailo(H)})$ is the probability that the
variable set $X_H$ assumes values $\textbf{0}$ in a kernel obtained from $p(x_V)$ by fixing
$X_{V \setminus C}$, and conditioning on $X_{\tailo(H)}$.  As a shorthand, we will denote the
parameter $q_C(X_{H} = \textbf{0} | x_{\tailo(H)})$ as $\theta_H(x_{\tailo(H)})$.

\begin{definition} Let $\nu : V\cup W \mapsto \{0,1\}$ be an assignment of values to the variables indexed by $V\cup W$.
Define $\nu(T)$ to be the values assigned to variables indexed by a subset $T\subseteq V\cup W$.
Let $\nu^{-1}(0) = \{ v \mid v\in V, \nu(v) = 0\}$.\par

A distribution $P(X_V \mid X_W)$ is said to be {\em  parameterized by} the set ${\mathfrak Q}_{\cal G}$, for CADMG $\cal G$ if:
\begin{eqnarray*}
p(X_V\!=\! \nu(V)\mid X_W\!=\! \nu(W) ) =
\!\!\!\!\!
\!\!\!\!\!
\!\!\!\!\!
\!\!\!\!\!
\displaystyle \sum_{B\;:\; \nu^{-1}(0)\cap V \subseteq B \subseteq V}
\!\!\!\!\!
\!\!\!\!\!
\!\!\!\!\!
(-1)^{\left| B\setminus \nu^{-1}(0)\right|} \times\\
\prod_{H\in \llbracket B\rrbracket_{\cal G}}
\theta_{H}(X_{\tailo(H)}= \nu (\tailo(H))),
\end{eqnarray*}
where the empty product is defined to be $1$, and $\llbracket B \rrbracket_{\cal G}$ is a partition of
nodes in $B$ given in \cite{shpitser11eid}.
\label{param}
\end{definition}

Note that this parameterization maps $\theta_H$ parameters to probabilities in a
CADMG via an inverse M\"obius transform.  Note also that this
parameterization generalizes both the standard Markov parameterization of DAGs
in terms of parameters of the form $p(x_i = 0 | \pa(x_i))$, and the 
parameterization of bidirected graph models given in \cite{drton08binary}.

\subsection{Example}

Consider an ADMG $\mathcal{G}$ shown in Fig. \ref{verma1} (b).
\begin{comment}
The following table
shows the intrinsic sets, recursive heads and tails in this graph:

\medskip

{\small
\begin{tabular}{c|ccccccccc}
$C$ & $\{X_1,X_2,X_3,X_4\}$ & $\{X_1,X_2,X_3\}$ & $\{X_2,X_3\}$ & $\{X_2\}$ \\
\hline
&\\[-10pt]
$H$ & $\{X_1,X_4\}$ & $\{X_1\}$ &  $\{X_2,X_3\}$ &  $\{X_2\}$ \\[4pt]
\hline
&\\[-10pt]
$T$ & $\{X_2,X_3\}$ & $\{X_2,X_3\}$ & $\emptyset$ & $\emptyset$ \\
\end{tabular}
\begin{tabular}{c|ccccccccc}
$C$ & $\{X_3\}$ & $\{X_4\}$ & $\{X_2,X_3,X_4\}$ \\
\hline
&\\[-10pt]
$H$ & $\{X_3\}$ & $\{X_4\}$ & $\{X_3,X_4\}$ \\[4pt]
\hline
&\\[-10pt]
$T$ & $\emptyset$ & $\{X_2\}$ & $\{ X_2 \}$ \\
\end{tabular}
}
\end{comment}
\noindent
The parameters associated with a binary model represented by this graph are:
$$\theta_{1,4}(x_2,x_3), \theta_{1}(x_2,x_3), \theta_{2,3}, \theta_{2}, \theta_{3}, \theta_{4}(x_2),
\theta_{3,4}(x_2)$$
Each of these parameters are functions which map binary values to
probabilities, which implies this binary model contains 15 parameters, in other
words it is saturated.  This is the case even though $\mathcal{G}$ is not
a complete graph.  A similar situation arises in mixed graph models of
conditional independence.  In such models a model represented by a graph with
missing edges may be saturated if nodes which are not direct neighbors are
connected by an \emph{inducing path} \cite{spirtes93causation}.
In particular, the mixed graph
shown in Fig. \ref{verma1} represents a saturated model of conditional
independence because there is an inducing path between $X_1$ and $X_4$.  The
reason this graph does not represent a saturated nested Markov model
is because truncations allow us to test independence of some pairs of
non-adjacent nodes, even if they are connected by an inducing path.  However,
there are some inducing paths which are ``dense'' enough such that truncations
cannot be used to test independence between node pairs connected by such
a path.  Such a dense inducing path exists in the graph in Fig. \ref{verma1} (b)
between $X_1$ and $X_4$.

As an illustration of our parameterization, for the graph in Fig. \ref{verma1} (b),
we have the following:
{\small
\begin{eqnarray*}
p(x_1=0,x_3=0,x_4 = 0, x_2 = 1) =\\
\theta_{1,4}(x_2 = 1, x_3 = 0) * \theta_{3}
- \theta_{1,4}(x_2 = 1, x_3 = 0) * \theta_{2,3}\\
p(x_1=0,x_3 = 0, x_2=1,x_4 = 1) =\\
\theta_{1}(x_2 = 1, x_3 = 0) * \theta_{3} 
- \theta_{1}(x_2 = 1, x_3 = 0) * \theta_{2,3} \\
- \theta_{1,4}(x_2 = 1, x_3 = 0) * \theta_{3} 
+ \theta_{1,4}(x_2 = 1, x_3 = 0) * \theta_{2,3}\\
\end{eqnarray*}
}
\vspace{-1.0cm}
\section{Parameter Fitting for Binary Nested Markov Models}

We now describe a parameter fitting algorithm based on the parameterization
in definition \ref{param}, which relates the parameters of an nested Markov
model and standard multinomial probabilities via the M\"obius inversion
formula.  We first describe this mapping in more detail.

For a given CADMG $\mathcal{G}$, 
define $\mathcal{P}_{\mathcal{G}}$ be the set of all multinomial probability
vectors in the simplex $\Delta_{2^{|V|}-1}$ which obey the nested factorization
according to $\mathcal{G}$, let $\mathcal{Q}_{\mathcal{G}}$ be the set of all vectors of
parameters which define coherent nested Markov models.

The mapping $\rho_{\mathcal{G}} : \mathcal{Q}_{\mathcal{G}} \mapsto
\mathcal{P}_{\mathcal{G}}$ in definition \ref{param}
can be viewed as a composition $\mu_{\mathcal{G}} \circ \tau_{\mathcal{G}}$ of
two mappings.  Here $\tau_{\mathcal{G}}$ maps $\mathcal{Q}_{\mathcal{G}}$ to
the set of all terms of the form
$\prod_{H \in \llbracket B \rrbracket_{\cal G}}
\theta_{H}(X_{\tailo(H)} = \nu(\tailo(H)))$ composed of parameters in
$\mathcal{Q}_{\mathcal{G}}$.  We denote this set by $\mathcal{T}_{\mathcal{G}}$.
The second mapping 
$\mu_{\mathcal{G}}$ maps $\mathcal{T}_{\mathcal{G}}$ to
$\mathcal{P}_{\mathcal{G}}$ via an inverse M\"obius transform.
In \cite{evans10maximum} these mappings were defined via element-wise matrix
operations, with
$\tau_{\mathcal{G}}$ defined via a matrix $P$ containing $0$ and $1$ entries,
and $\mu_{\mathcal{G}}$ defined via a matrix $M$ containing $0,1,-1$ entries.
There may be more efficient representations of these mappings.  In
particular $\mu_{\mathcal{G}}$ may be evaluated via the fast M\"obius transform
\cite{kennes91moebius}.  Such an efficient mapping was given in
\cite{shpitser11eid}.

Note that $\rho_{\mathcal{G}}$ is smooth with respect to each
parameter.  This implies we can solve many optimization
problems for functions expressed in terms of $\rho_{\mathcal{G}}$ using
standard iterative methods.  The difficulty is that the fitting algorithm must
be defined in such a way that each step that starts in the parameter space
$\mathcal{Q}_{\mathcal{G}}$ stays in $\mathcal{Q}_{\mathcal{G}}$.  We use the
approach taken in \cite{evans10maximum}, where a single step of the fitting
algorithm updates
the estimates for all and only parameters which refer to a particular
vertex $v \in V$.  For a particular vector
$\textbf{q} \in \mathcal{Q}_{\mathcal{G}}$, let $\textbf{q}(v)$ be the set of
parameters whose heads contain $v$.  Let $\mathcal{Q}_{\mathcal{G}}(v)$ be the
subset of $\mathcal{Q}_{\mathcal{G}}$ containing only such vectors.  Then
the restriction of $\rho_{\mathcal{G}}$ to $\mathcal{Q}_{\mathcal{G}}(v)$ is a
linear function since any such parameter occurs at most once in in a term
in $\mathcal{T}_{\mathcal{G}}$.  This implies that 
$\rho_{\mathcal{G}}$ can be expressed as $A_v * \mathcal{Q}_{\mathcal{G}}(v) -
\textbf{b}_v$ for some matrix $A_v$ and vector $\textbf{b}_v$.  To remain within
$\mathcal{Q}_{\mathcal{G}}$ it suffices to maintain the constraint that
$A_v\textbf{q}(v) \geq \textbf{b}_v$.

\subsection{Maximum Likelihood Parameter Fitting}

We are now ready to describe our parameter fitting algorithm.  Our scheme
closely follows that in \cite{evans10maximum}, albeit with a different
parameterization.
The algorithm iteratively updates parameters $\textbf{q}(v)$ for every vertex
$v$ in turn, and at each step maximizes the log likelihood via gradient
ascent.  For the purposes of this paper, we assume strictly positive counts in
our data.  The case of zero counts gives rise to certain statistical
complications, and will be handled in subsequent work.


For a particular vertex $v$, the function we are optimizing has the form
$\log \mathcal{L}_{\mathcal{G}}(\textbf{q}(v)) =
	\sum_{i} n_i \log \rho_{\mathcal{G}}(\textbf{q}(v))$
where $\rho_{\mathcal{G}}$ is restricted to $\mathcal{Q}_{\mathcal{G}}(v)$ and
is thus a linear function in $\textbf{q}(v)$.

Our fitting algorithm is given in Fig. \ref{fitting}.  Our choice of
$L$ is the log likelihood function which is strictly concave in
$\textbf{q}(v)$ by above, while our initial guess for $\textbf{q}$ are the
parameters which define a fully independent model.  The optimization problem in
line 2 can be solved by standard gradient ascent methods.

%

\begin{figure}
\textbf{Q-FIT}($\mathcal{G}, \textbf{q}, L(\rho_{\mathcal{G}})$)\\
INPUT: $\textbf{G}$ an ADMG, \textbf{q} a set of parameters defining a model
	which obeys the nested factorization wrt $\mathcal{G}, L(\rho_{\mathcal{G}})$
	a concave function defined in terms of $\rho_{\mathcal{G}}$.\\
OUTPUT: $\hat{\textbf{q}}$, a local maximum in the surface defined
on $\mathcal{Q}_{\mathcal{G}}$ via $L(\rho_{\mathcal{G}})$.\\
\vspace{-0.5cm}
\begin{itemize}
\item[] Cycle through each vertex $v$ in $\mathcal{G}$, and do
	\begin{itemize}
	\item[1] Construct the constraint matrices $A_v, \textbf{b}_v$.
	\item[2] Fit $\textbf{q}(v)$ to obtain new estimate $\textbf{q}^*$
		maximizing $L(\rho_{\mathcal{G}})$ subject
		to $A_v \textbf{q}(v) \geq \textbf{b}_v$.
	\item[3] If $\textbf{q}^*$ sufficiently close to $\textbf{q}$, return
		$\textbf{q}^*$.
	\item[4] Otherwise, set $\textbf{q}$ to $\textbf{q}^*$.
	\end{itemize}
\end{itemize}
\caption{A parameter fitting algorithm for nested Markov models.}
\label{fitting}
\end{figure}

\section{Structure Learning in Nested Markov Models}

A fitting algorithm which maximizes likelihood allows us to do structure
learning in nested Markov models, using standard search and score
methods which use likelihood-based scoring criteria such as BIC
\cite{schwarz78bic}.

The algorithm is a standard greedy local search augmented with a tabu
meta-heuristic.  We found a meta-heuristic necessary for our search procedure
because a complete theory of equivalence of models with respect to
post-truncation independences is not yet available.  Because we do not yet
understand equivalence in this setting, we are unable to define efficient local
steps which always move across equivalence classes as in the GES algorithm
for DAGs \cite{chickering02ges}.  Without such steps, in order to achieve
reasonable local
minima in the score surface, the algorithm must be able to move across
score plateaus.

\smallskip

For the purposes of our experiments, we used the BIC scoring
function, although our approach does not require this, and any competing
scoring method could have been used.  We chose BIC due to its desirable
asymptotic properties.

\smallskip

We interpret the output of our search procedure to be the ``best'' mixed graph
model under the assumption that every post-truncation independence
observed in the data has a structural explanation.  This assumption is a
natural generalization of the faithfulness \cite{spirtes93causation}, or
stability \cite{pearl00causality} assumption from the conditional
independence setting to the post-truncation independence setting.  We do not
pursue the precise statement of this assumption in this paper, since doing so
entails defining a strong global Markov property for nested Markov models (the
post-truncation analogue of d-separation in DAGs and m-separation in
mixed graphs).  This property is sufficiently intricate that its definition
and properties are developed in a companion paper.

\subsection{Implementation}

We implemented fitting and search algorithms using the R language
\cite{r04book}.  Our implementation was based on an older implementation
of fitting and search for mixed graph models of conditional independence
\cite{evans10maximum}.

\section{Experiments}

To illustrate our search and score method, we used 
simulated data.

\subsection{Simulated Data from DAG Marginal Models With Post-Truncation
Constraints}

\begin{figure}
\begin{center}
  \begin{tikzpicture}[>=stealth, node distance=1.0cm]

  \begin{scope}
    \tikzstyle{format} = [draw, thick, rectangle, minimum size=6mm, rounded
        corners=3mm]
    \path[->]	node[format] (x1) {$x_1$}
    		node[format, right of=x1] (x2) {$x_2$}
                  (x1) edge (x2)
		node[format, right of=x2] (x3) {$x_3$}
                  (x2) edge (x3)
		node[format, right of=x3] (x4) {$x_4$}
                  (x3) edge (x4)
		node[format, above of=x3] (u) {$u$}
		  (u) edge (x2)
		  (u) edge (x4)
		;

	\path[->] node[below of=x3] (label) {$(a)$};
  \end{scope}

  \begin{scope}[xshift = 4.0cm]
    \tikzstyle{format} = [draw, thick, rectangle, minimum size=6mm, rounded
        corners=3mm]
    \path[->]	node[format] (x1) {$x_1$}
    		node[format, right of=x1] (x2) {$x_2$}
                  (x1) edge (x2)
		node[format, right of=x2] (x3) {$x_3$}
                  (x2) edge (x3)
		node[format, right of=x3] (x4) {$x_4$}
                  (x3) edge (x4)
		node[format, above of=x3] (x5) {$x_5$}
		node[format, above of=x2] (u1) {$u_1$}
			(u1) edge (x2)
			(u1) edge (x5)
		node[format, above of=x4] (u2) {$u_2$}
			(u2) edge (x5)
			(u2) edge (x4)
		;

	\path[->] node[below of=x3] (label) {$(b)$};
  \end{scope}

  \end{tikzpicture}
\end{center}
\caption{DAG models used in our simulation experiments.}
\label{simulations}
\end{figure}
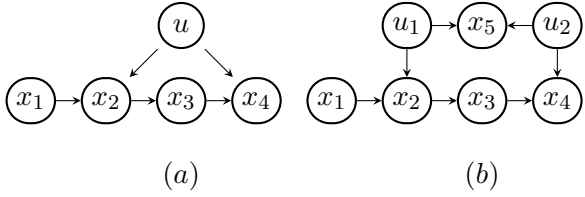

To demonstrate that our algorithm can successfully learn ``interesting''
graphs distinct from known DAG and MAG equivalence classes, we have used
search and score on simulated data obtained from DAG models with latent
variables shown in Fig. \ref{simulations}.  Both of these models
are known to contain post-truncation independences.  The observable nodes
$X_1$, $X_2$, $X_3$, $X_4$ in the graph shown in Fig. \ref{simulations} (a)
correspond to binary random variables, while the latent node $U$ corresponds
to a discrete random variable with 16 possible values.  Similarly, the
observable nodes $X_1$, $X_2$, $X_3$, $X_4$, $X_5$ in the graph shown in Fig.
\ref{simulations} (b) correspond to binary random variables,  while the
latent nodes $U_1$, $U_2$ correspond to discrete random variables with 8
possible values.

The model shown in the graph in Fig. \ref{simulations} (a)
contains two independence constraints over observable variables.  The first is
an ordinary conditional independence constraint $(X_1 \ci X_3 | X_2)$.  The
second is a post-truncation independence which states that $X_1 \ci X_4$ after
truncating out $P(x_3 \mid x_2)$.  The model shown in the graph in
Fig. \ref{simulations} (b) contains three independence constraints over
observable variables.
The first two are ordinary independence constraints which state that
$(X_3 \ci X_1 | X_2)$ and $(X_4 \ci X_1,X_2 | X_3)$.  The last is a
post-truncation independence which states that $X_4,X_5 \ci X_1$ after
truncating out $P(x_3 \mid x_2)$.  Note that $X_4$ and $X_5$ can both
be made conditionally independent of $X_1$, but not by using the same
conditioning set.

\subsection{Results}

We chose parameters of the DAG models shown in Fig. \ref{simulations} in such
a way as to ensure ``approximately faithful'' models.
We then generated samples from our models and retained the values
of only $X_1, X_2, X_3, X_4$ in the first model, and of only
$X_1, X_2, X_3, X_4, X_5$ in the second model.
We evaluated the performance of our structure learning algorithm on datasets
ranging from 500 to 5000 samples (in 500 sample increments).  For each dataset
size, we generated 1000 datasets randomly from the true models.

\begin{figure*}[ht]
\begin{minipage}[b]{0.5\linewidth}
\centering
\scalebox{0.37}{
\includegraphics[bb = 4 16 477 480]{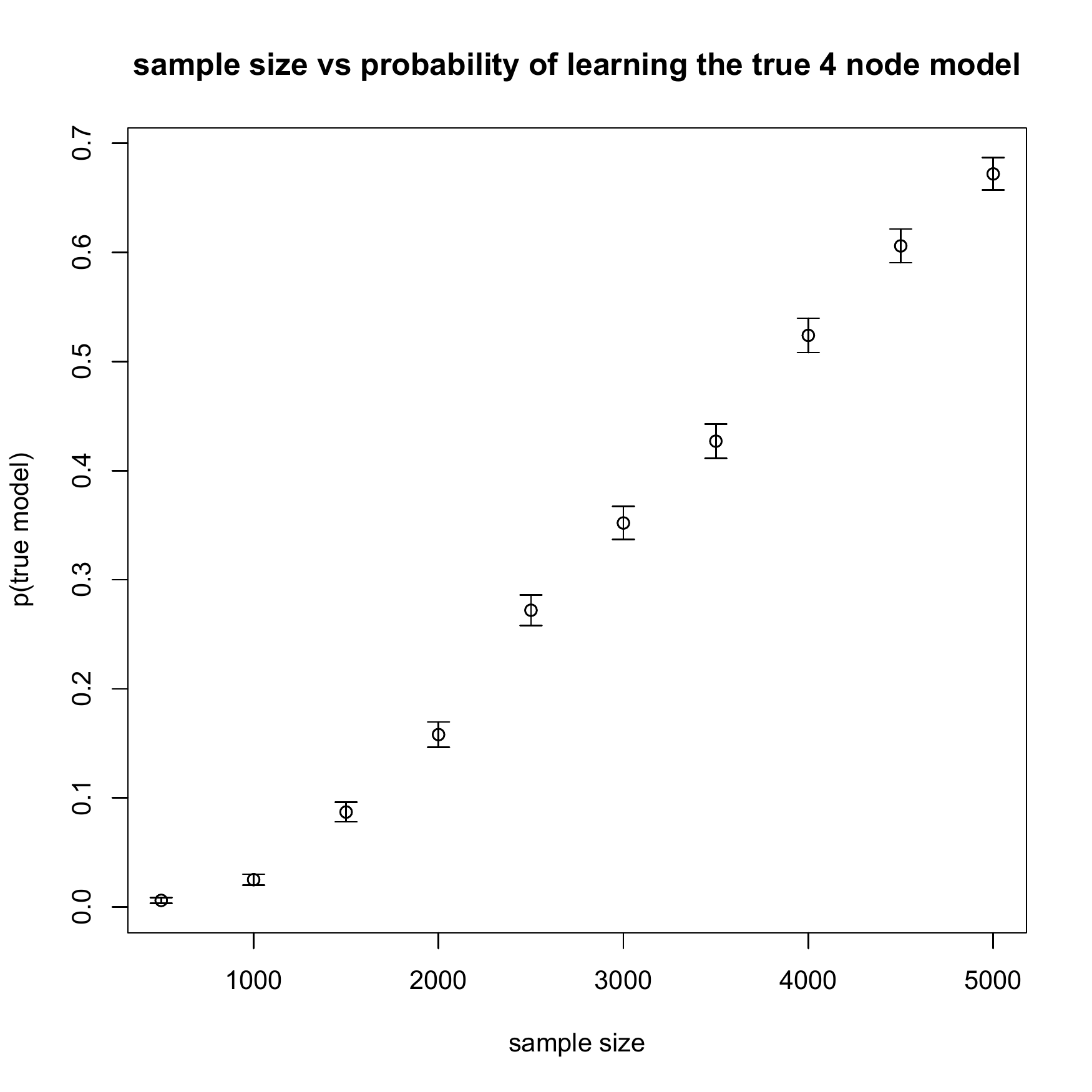}
}
\end{minipage}
\hspace{0.5cm}
\begin{minipage}[b]{0.5\linewidth}
\centering
\scalebox{0.37}{
\includegraphics[bb = 4 16 477 480]{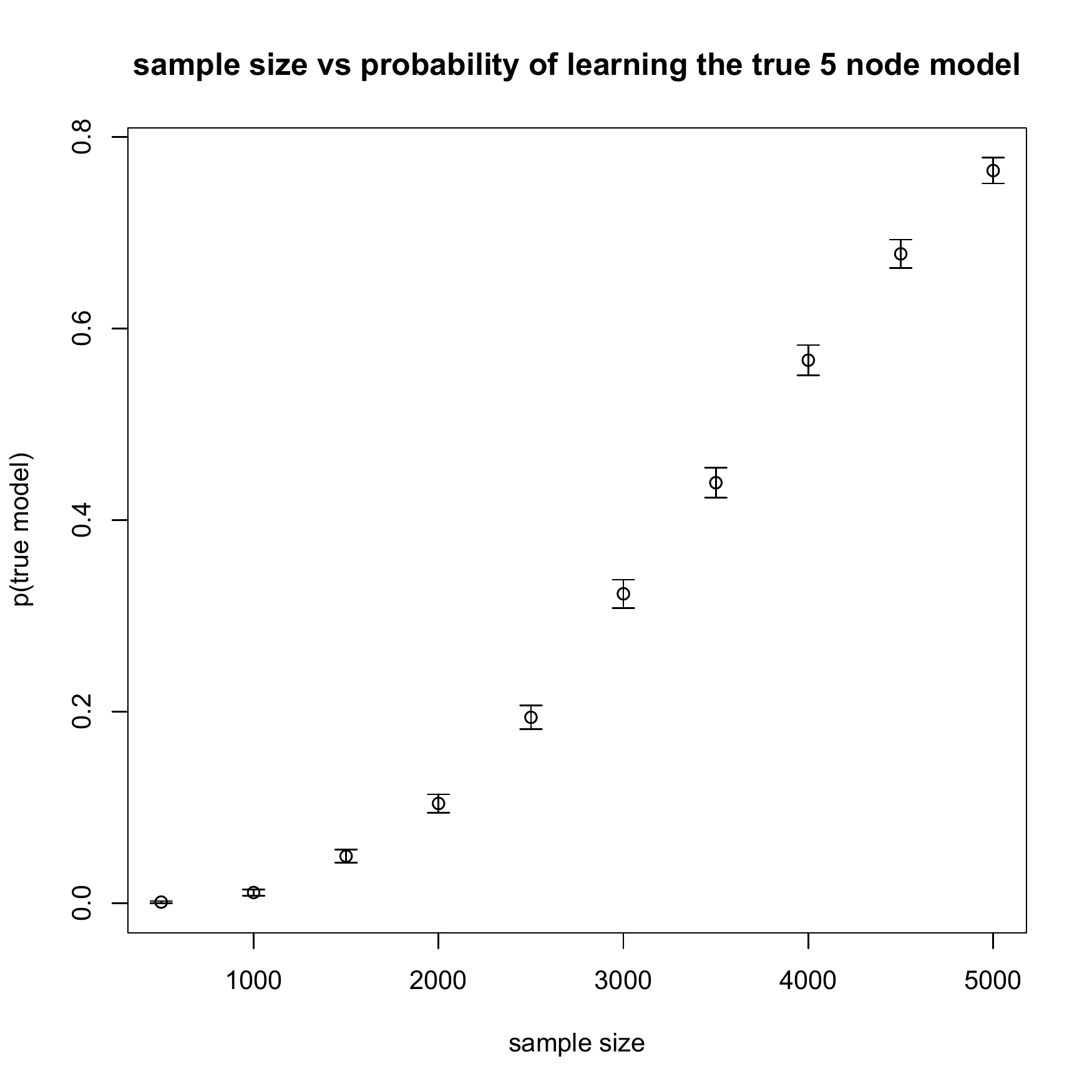}
}
\end{minipage}
\caption[]{(a) Probability of learning the true 4 node model vs sample size.
(b) Probability of learning the true 5 node model vs sample size.}
\label{plots}
\end{figure*}

Figures \ref{plots} (a) and (b) show our results.  The probability of
learning the true model grows linearly with sample size, with 672 4 node
models out of 1000 correctly recovered, and 765 5 node models out of 1000
correctly recovered, from 5000 sample datasets.  By ``correctly recovered''
we mean that our search procedure returned one of the graphs shown in
Fig. \ref{type_d} in datasets obtained from a DAG model in
Fig. \ref{simulations} (a), and one of the graphs in the appropriate equivalence class
in datasets obtained from a DAG model in Fig. \ref{simulations} (b).
Although
there is currently no complete theory of observational equivalence of
nested Markov models, as there is for Markov factorizing models, we do provide
evidence for the characterization of all equivalence classes of 4 node models in
the next section.

\section{Equivalence Conjecture for Mixed Graphs of Four Nodes}

When describing the search and score algorithm for nested Markov models, we
mentioned that no complete theory of equivalence of models with respect to
post-truncation independence currently exists.  Characterization of
equivalence does exist for DAGs \cite{verma90equiv},
and MAGs \cite{ali09equiv}.
In this section, we present an investigation of the issue of equivalence with
respect to post-truncation independence for the special case of mixed graphs
of four nodes.

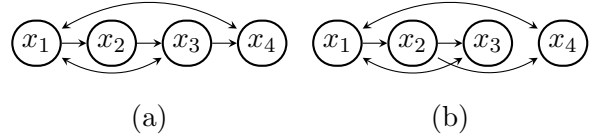
\begin{figure}
\begin{center}
  \begin{tikzpicture}[>=stealth, node distance=1.0cm]

  \begin{scope}
    \tikzstyle{format} = [draw, thick, rectangle, minimum size=6mm, rounded
        corners=3mm]
	\path[->]
		node[format] (x1) {$x_1$}
		node[format, right of=x1] (x2) {$x_2$}
			(x1) edge (x2)
		node[format, right of=x2] (x3) {$x_3$}
			(x2) edge (x3)
			(x1) edge[<->, bend right] (x3)
		node[format, right of=x3] (x4) {$x_4$}
			(x3) edge (x4)
			(x1) edge[<->, bend left] (x4)
		node[below of=x2,xshift=0.5cm] (label) {(a)};
  \end{scope}

  \begin{scope}
    \tikzstyle{format} = [draw, thick, rectangle, minimum size=6mm, rounded
        corners=3mm]
	\path[->, xshift = 4.0cm]
		node[format] (x1) {$x_1$}
		node[format, right of=x1] (x2) {$x_2$}
			(x1) edge (x2)
		node[format, right of=x2] (x3) {$x_3$}
			(x2) edge (x3)
			(x1) edge[<->, bend right] (x3)
		node[format, right of=x3] (x4) {$x_4$}
			(x2) edge[bend right] (x4)
			(x1) edge[<->, bend left] (x4)
		node[below of=x2, xshift=0.5cm] (label) {(b)};
  \end{scope}

  \end{tikzpicture}
\end{center}
\caption{(a) An equivalence class pattern containing 24 equivalence classes
which in turn contain 1 graph each.  (b) An equivalence class pattern
containing 12 equivalence classes which in turn contain 1 graph each.}
\label{type_ab}
\end{figure}

It is known that there are exactly 543 four node DAGs
(sequence A003024 in OEIS).
This implies that there are $543 * 2^6 = 34752$ ADMGs with four nodes.  These
ADMGs are arranged into 185 equivalence classes representing DAG models of
conditional independence, and 63 equivalence classes representing models of
independence which can be represented by a mixed graph, but not a DAG, for
a total of 248 equivalence classes.  If we consider nested Markov models, which
in addition to conditional independences imply post-truncation independences,
we expect the number of equivalence classes to expand, since two mixed graphs
may agree on all conditional independences, but disagree on post-truncation
independences.  For instance, this is the case for the ADMG shown in
Fig. \ref{type_ab} (a) and a complete DAG.

We conjecture that for a given ADMG $\mathcal{G}$, if the model
of conditional independence \cite{richardson09factorization} and
a nested Markov model agree on the parameter count, then they define the same
model \cite{richardson11generalized}.  This conjecture would
imply that it is sufficient to characterize equivalence in ADMGs where the model in
\cite{richardson09factorization} and the nested Markov model give different
parameter counts.  There are exactly 228 such ADMGs.

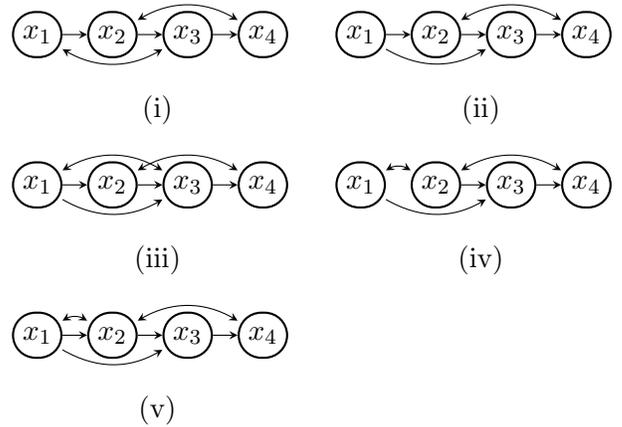
\begin{figure}
\begin{center}
  \begin{tikzpicture}[>=stealth, node distance=1.0cm]

  \begin{scope}
    \tikzstyle{format} = [draw, thick, rectangle, minimum size=6mm, rounded
        corners=3mm]
	\path[->]
		node[format] (x1) {$x_1$}
		node[format, right of=x1] (x2) {$x_2$}
			(x1) edge (x2)
		node[format, right of=x2] (x3) {$x_3$}
			(x2) edge (x3)
			(x1) edge[<->, bend right] (x3)
		node[format, right of=x3] (x4) {$x_4$}
			(x3) edge (x4)
			(x2) edge[<->, bend left] (x4)
		node[below of=x2, xshift=0.6cm] (label) {(i)};
  \end{scope}

  \begin{scope}
    \tikzstyle{format} = [draw, thick, rectangle, minimum size=6mm, rounded
        corners=3mm]
	\path[->, xshift = 4.3cm]
		node[format] (x1) {$x_1$}
		node[format, right of=x1] (x2) {$x_2$}
			(x1) edge (x2)
		node[format, right of=x2] (x3) {$x_3$}
			(x2) edge (x3)
			(x1) edge[bend right] (x3)
		node[format, right of=x3] (x4) {$x_4$}
			(x3) edge (x4)
			(x2) edge[<->, bend left] (x4)
		node[below of=x2, xshift=0.6cm] (label) {(ii)};
  \end{scope}

  \begin{scope}
    \tikzstyle{format} = [draw, thick, rectangle, minimum size=6mm, rounded
        corners=3mm]
	\path[->, yshift = -2.0cm]
		node[format] (x1) {$x_1$}
		node[format, right of=x1] (x2) {$x_2$}
			(x1) edge (x2)
		node[format, right of=x2] (x3) {$x_3$}
			(x2) edge (x3)
			(x1) edge[bend right] (x3)
			(x1) edge[<->, bend left] (x3)
		node[format, right of=x3] (x4) {$x_4$}
			(x3) edge (x4)
			(x2) edge[<->, bend left] (x4)
		node[below of=x2, xshift=0.6cm] (label) {(iii)};
  \end{scope}

  \begin{scope}
    \tikzstyle{format} = [draw, thick, rectangle, minimum size=6mm, rounded
        corners=3mm]
	\path[->, xshift = 4.3cm, yshift = -2.0cm]
		node[format] (x1) {$x_1$}
		node[format, right of=x1] (x2) {$x_2$}
			(x1) edge[<->, bend left] (x2)
		node[format, right of=x2] (x3) {$x_3$}
			(x2) edge (x3)
			(x1) edge[bend right] (x3)
		node[format, right of=x3] (x4) {$x_4$}
			(x3) edge (x4)
			(x2) edge[<->, bend left] (x4)
		node[below of=x2, xshift=0.6cm] (label) {(iv)};
  \end{scope}

  \begin{scope}
    \tikzstyle{format} = [draw, thick, rectangle, minimum size=6mm, rounded
        corners=3mm]
	\path[->, yshift = -4.0cm]
		node[format] (x1) {$x_1$}
		node[format, right of=x1] (x2) {$x_2$}
			(x1) edge (x2)
			(x1) edge[<->, bend left] (x2)
		node[format, right of=x2] (x3) {$x_3$}
			(x2) edge (x3)
			(x1) edge[bend right] (x3)
		node[format, right of=x3] (x4) {$x_4$}
			(x3) edge (x4)
			(x2) edge[<->, bend left] (x4)
		node[below of=x2, xshift=0.6cm] (label) {(v)};
  \end{scope}

  \end{tikzpicture}
\end{center}
\caption{(i)-(v) Five graph patterns together representing 24
equivalence classes, each class containing 5 graphs, one from each pattern.}
\label{type_c}
\end{figure}

\begin{figure}
\begin{center}
  \begin{tikzpicture}[>=stealth, node distance=1.0cm]

  \begin{scope}
    \tikzstyle{format} = [draw, thick, rectangle, minimum size=6mm, rounded
        corners=3mm]
	\path[->]
		node[format] (x1) {$x_1$}
		node[format, right of=x1] (x2) {$x_2$}
			(x1) edge (x2)
		node[format, right of=x2] (x3) {$x_3$}
			(x2) edge (x3)
		node[format, right of=x3] (x4) {$x_4$}
			(x3) edge (x4)
			(x2) edge[<->, bend left] (x4)
		node[below of=x2,xshift=0.6cm] (label) {(i)};
  \end{scope}

  \begin{scope}
    \tikzstyle{format} = [draw, thick, rectangle, minimum size=6mm, rounded
        corners=3mm]
	\path[->, xshift = 4.3cm]
		node[format] (x1) {$x_1$}
		node[format, right of=x1] (x2) {$x_2$}
			(x1) edge[<->, bend left] (x2)
		node[format, right of=x2] (x3) {$x_3$}
			(x2) edge (x3)
		node[format, right of=x3] (x4) {$x_4$}
			(x3) edge (x4)
			(x2) edge[<->, bend left] (x4)
		node[below of=x2,xshift=0.6cm] (label) {(ii)};
  \end{scope}

  \begin{scope}
    \tikzstyle{format} = [draw, thick, rectangle, minimum size=6mm, rounded
        corners=3mm]
	\path[->, yshift = -2.0cm]
		node[format] (x1) {$x_1$}
		node[format, right of=x1] (x2) {$x_2$}
			(x1) edge (x2)
			(x1) edge[<->, bend left] (x2)
		node[format, right of=x2] (x3) {$x_3$}
			(x2) edge (x3)
		node[format, right of=x3] (x4) {$x_4$}
			(x3) edge (x4)
			(x2) edge[<->, bend left] (x4)
		node[below of=x2,xshift=0.6cm] (label) {(iii)};
  \end{scope}

  \end{tikzpicture}
\end{center}
\caption{(i)-(iii) Three graph patterns together representing 24
equivalence classes, each class containing 3 graphs, one from each pattern.}
\label{type_d}
\end{figure}
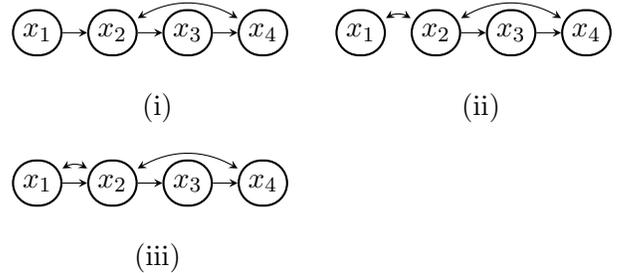

We conjecture that these 228 ADMGs are arranged in 84 equivalence classes.
These 84 classes fall in a small number of graph
patterns with multiple classes having the same pattern but different vertex
labeling.  Specifically, of the 84 classes, 24 are of type (a), shown in
Fig. \ref{type_ab} (a), 12 are of type (b), shown in Fig. \ref{type_ab} (b),
24 contain graphs with the patterns shown in Fig. \ref{type_c}, and
24 contain graphs with the patterns shown in Fig. \ref{type_d}.

The model contained in one of 24 singleton equivalence classes shown in
Fig. \ref{type_ab} (a) has a single post-truncation independence which
states (up to node relabeling) that
$X_4 \ci X_2 | X_3$ in the distribution obtained from $P(x_1,x_2,x_3,x_4)$
after truncating out $P(x_3 \mid x_2,x_1)$.

The model contained in one of 12 singleton equivalence classes shown in
Fig. \ref{type_ab} (b) has a single post-truncation independence which
states (up to node relabeling) that
$X_4 \ci X_3 | X_2$ in the distribution obtained from $P(x_1,x_2,x_3,x_4)$
after truncating out $P(x_2 \mid x_1)$.
The advantage of exploiting post-truncation independence is clear in the
cases shown in Fig. \ref{type_ab}.  If the best scoring model lies in these
classes, then we can recover the model structure exactly just from observing
a single post-truncation independence, whereas if we restricted ourselves to
conditional independence we would be unable to distinguish models in these
classes from saturated models.

The model contained in one of 24 equivalence classes shown in
Fig. \ref{type_c} (which contains 5 ADMGs) has a single post-truncation
independence which states (up to node relabeling) that
$X_4 \ci X_1$ in the distribution obtained from $P(x_1,x_2,x_3,x_4)$ after
truncating out $P(x_3 \mid x_2,x_1)$.
In the case of models shown in Fig. \ref{type_c}, even though the equivalence
class contains 5 graphs, these graphs agree on many interesting (from a causal
point of view) structural features.  In particular, in all elements of a
particular class the
following edges are present (up to node relabeling):
$X_3 \to X_4, X_2 \to X_3, X_2 \leftrightarrow X_4$.
As before, these models are indistinguishable from the saturated model with
respect to standard conditional independence constraints.

The model contained in one of 24 equivalence classes shown in
Fig. \ref{type_d} (which contains 3 ADMGs) has one conditional independence
which states (up to node relabeling) that $X_3 \ci X_1 | X_2$, and one
post-truncation independence which states (up to node relabeling) that
$X_4 \ci X_1$ in the distribution obtained from $P(x_1,x_2,x_3,x_4)$ after
truncating out $P(x_3 \mid x_2)$.
Similarly, models shown in Fig. \ref{type_c} are members of equivalence
classes contains 3 graphs, yet these graphs agree on many interesting
structural features.  As before, in all elements of a particular class the
following edges are present (up to node relabeling):
$X_3 \to X_4, X_2 \to X_3, X_2 \leftrightarrow X_4$.
These models are indistinguishable from the (DAG) model asserting a single
conditional independence $X_3 \ci X_1 | X_2$, with respect to standard
conditional independence constraints.

To confirm our conjecture, we have verified that log-likelihood values of
nested Markov models obtained by $\textbf{Q-FIT}$ from datasets generated
from a four node saturated model
are always the same within our conjectured classes.

Finally, we note that if our conjecture is correct, post-truncation
independences occur in about $25\%$ ($84/(248+84)$) of four node mixed graph
equivalence classes.  This
suggests that, far from being ``rare and exotic,'' these constraints may be
fairly common in latent variable models.  This is particularly encouraging
since post-truncation constraints seem to be quite informative for causal
discovery.

\section{Discussion}


We described a new class of graphical models called nested Markov models,
 which can be viewed as the ``closure'' of DAG marginal models which preserves
all equality constraints.  These constraints include standard conditional
independence constraints, and less well-understood constraints which manifest
after a truncation operation, which corresponds to dividing by a conditional
distribution.  We have given a nested factorization of these models which
generalizes the standard Markov factorization of DAG models, and the
factorization of bidirected graph models \cite{drton08binary}.
We have given a parameterization for discrete models, and used this
parameterization to give a parameter fitting and structure learning algorithm
for nested Markov models.  Together with results in \cite{shpitser11eid},
our parameter fitting scheme gives an MLE for any identifiable causal effect
in discrete nested Markov models.

We have applied our structure learning algorithm
to simulated data.
We have shown that our algorithm can correctly distinguish models based on
post-truncation independences, which no other currently known discovery
algorithm is capable of doing.  Finally, we used our fitting procedure to
justify a conjecture which characterizes model equivalence with respect to
post-truncation independence in four node mixed graph models.

The advantage of our approach is twofold.  First, by representing
latent variables implicitly, we are able to reason over a potentially
infinite set of DAG models which can give rise to a particular pattern of
constraints.  Second, our machinery explicitly incorporates post-truncation
independence, which is a kind of equality
constraint which generalizes conditional independence, and
which can be used to distinguish models which are not distinguishable
with respect to standard conditional independence.  We have shown cases where
discovering a single post-truncation independence is sufficient to recover
the full structure of an ADMG without any ambiguity, though the corresponding
model has no standard conditional independence constraints.

Both our nested factorization and the post-truncation independences this
factorization implies have an intuitive causal interpretation.  The
factorization can be thought of as decomposing the joint distribution into
tractable pieces corresponding to joint direct effects on bidirected connected
sets, while the post-truncation independence correspond to (identifiable)
dormant independence constraints \cite{shpitser08dormant},
which can be viewed as either an absence of
some direct effect, or a decomposition of a joint direct effect into multiple
smaller joint direct effects.

{
\small
\bibliographystyle{plain}
\bibliography{references}

\begin{thebibliography}{10}

\bibitem{ali09equiv}
Ayesha Ali, Thomas~S. Richardson, and Peter Spirtes.
\newblock Markov equivalence for ancestral graphs.
\newblock {\em Annals of Statistics}, 37:2808--2837, 2009.

\bibitem{chickering02ges}
David~Maxwell Chickering.
\newblock Optimal structure identifiation with greedy search.
\newblock {\em Journal of Machine Learning Research}, 3:507--554, 2002.

\bibitem{drton08binary}
M.~Drton and T.S. Richardson.
\newblock Binary models for marginal independence.
\newblock {\em J. Roy. Statist. Soc. Ser. B}, 70(2):287--309, 2008.

\bibitem{evans10maximum}
Robin~J. Evans and Thomas~S. Richardson.
\newblock Maximum likelihood fitting of acyclic directed mixed graphs to binary
  data.
\newblock In {\em Proceedings of the Twenty Sixth Conference on Uncertainty in
  Artificial Intelligence}, volume~26, 2010.

\bibitem{kennes91moebius}
R.~Kennes.
\newblock Computational aspects of the moebius transform of a graph.
\newblock {\em IEEE Transactions on Systems, Man, and Cybernetics},
  22:201--223, 1991.

\bibitem{lauritzen96graphical}
S.L. Lauritzen.
\newblock {\em Graphical Models}.
\newblock Oxford, U.K.: Clarendon, 1996.

\bibitem{pearl00causality}
Judea Pearl.
\newblock {\em Causality: Models, Reasoning, and Inference}.
\newblock Cambridge University Press, 2000.

\bibitem{r04book}
{R Development Core Team}.
\newblock {\em R: A language and environment for statistical computing}.
\newblock Vienna: R Foundation for Statistical Computing, 2004.

\bibitem{richardson09factorization}
Thomas~S. Richardson.
\newblock A factorization criterion for acyclic directed mixed graphs.
\newblock In {\em 25th Conference on Uncertainty in Artificial Intelligence},
  2009.

\bibitem{richardson11generalized}
Thomas~S. Richardson, James~M. Robins, and Ilya Shpitser.
\newblock Generalized {M}arkov models associated with acyclic directed mixed
  graphs.
\newblock Unpublished manuscript, 2011.

\bibitem{richardson12nested}
Thomas~S. Richardson, James~M. Robins, and Ilya Shpitser.
\newblock Nested markov properties for acyclic directed mixed graphs.
\newblock In {\em 28th Conference on Uncertainty in Artificial Intelligence
  (UAI-12)}. AUAI Press, 2012.

\bibitem{robins:1999}
James~M. Robins.
\newblock Testing and estimation of direct effects by reparameterizing directed
  acyclic graphs with structural nested models.
\newblock In C.~Glymour and G.~Cooper, editors, {\em Computation, Causation,
  and Discovery}, pages 349--405. MIT Press, Cambridge, MA, 1999.

\bibitem{robins86new}
J.M. Robins.
\newblock A new approach to causal inference in mortality studies with
  sustained exposure periods -- application to control of the healthy worker
  survivor effect.
\newblock {\em Mathematical Modeling}, 7:1393--1512, 1986.

\bibitem{schwarz78bic}
Gideon~E. Schwarz.
\newblock Estimating the dimension of a model.
\newblock {\em Annals of Statistics}, 6:461--464, 1978.

\bibitem{shpitser08dormant}
Ilya Shpitser and Judea Pearl.
\newblock Dormant independence.
\newblock Technical Report R-340, Cognitive Systems Laboratory, University of
  California, Los Angeles, 2008.

\bibitem{shpitser09edge}
Ilya Shpitser, Thomas~S. Richardson, and James~M. Robins.
\newblock Testing edges by truncations.
\newblock In {\em International Joint Conference on Artificial Intelligence},
  volume~21, pages 1957--1963, 2009.

\bibitem{shpitser11eid}
Ilya Shpitser, Thomas~S. Richardson, and James~M. Robins.
\newblock An efficient algorithm for computing interventional distributions in
  latent variable causal models.
\newblock In {\em 27th Conference on Uncertainty in Artificial Intelligence
  (UAI-11)}. AUAI Press, 2011.

\bibitem{spirtes93causation}
P.~Spirtes, C.~Glymour, and R.~Scheines.
\newblock {\em Causation, Prediction, and Search}.
\newblock Springer Verlag, New York, 1993.

\bibitem{verma90equiv}
T.~S. Verma and Judea Pearl.
\newblock Equivalence and synthesis of causal models.
\newblock Technical Report R-150, Department of Computer Science, University of
  California, Los Angeles, 1990.

\end{thebibliography}
}

\end{document}